\documentclass[runningheads]{llncs}
\usepackage{eccv}

\usepackage{eccvabbrv}

\usepackage{graphicx}
\usepackage{booktabs}

\usepackage[accsupp]{axessibility}  
\usepackage{multirow}
\usepackage{bm}
\usepackage{booktabs}
\usepackage{makecell}
\usepackage{bigstrut}
\usepackage{subcaption}
\usepackage[table]{xcolor}
\usepackage[utf8]{inputenc}
\usepackage{fontawesome5}
\usepackage{xcolor}
\usepackage{caption}
\usepackage{graphicx}
\usepackage{amssymb}
\usepackage{setspace}
\usepackage{threeparttable}
\usepackage[font=small,labelfont=bf,labelsep=space]{caption}
\usepackage[font=scriptsize,labelfont=bf]{subcaption}
\usepackage{pifont}
\usepackage{tabularx} 
\usepackage{algorithm}
\usepackage{algorithmic}

\definecolor{blue}{rgb}{0.22, 0.22, 0.95}
\definecolor{bblue}{rgb}{0.12, 0.43, 0.84}
\definecolor{gray}{rgb}{0.29, 0.31, 0.31}
\definecolor{green}{rgb}{0.22, 0.7, 0.22}
\definecolor{lred}{rgb}{0.85, 0.27, 0.08}
\definecolor{tred}{rgb}{0.459,0.184,0.063}
\definecolor{orange}{rgb}{1.0, 0.4, 0}

\definecolor{argosNavy}{HTML}{0C2340}
\definecolor{argosWho}{HTML}{2776B8}
\definecolor{argosWhere}{HTML}{33B4A3}
\definecolor{argosWhen}{HTML}{E87838}

\newcommand\blfootnote[1]{%
  \begingroup
  \renewcommand\thefootnote{}%
  \renewcommand\thempfn{}%
  \footnotetext{#1}%
  \endgroup
}

\usepackage[pagebackref]{hyperref}

\usepackage{orcidlink}

\begin{document}

\title{ARGOS: Who, Where, and When\\ in Agentic Multi-Camera Person Search}

\author{Myungchul Kim\inst{1}\orcidlink{0000-0003-2959-7405} \and
Kwanyong Park\inst{2}$^{\dagger}$\orcidlink{0000-0002-6198-1610} \and
Junmo Kim\inst{1}$^{\dagger}$\orcidlink{0000-0002-7174-7932} \and
In So Kweon\inst{1,3}$^{\dagger}$\orcidlink{0000-0001-9626-5983}}

\authorrunning{M.~Kim et al.}
\titlerunning{ARGOS: Agentic Multi-Camera Person Search}

\institute{Korea Advanced Institute of Science and Technology (KAIST), Republic of Korea
\and
University of Seoul, Republic of Korea
\and
Korea Institute of Science and Technology (KIST), Republic of Korea\\
\email{gritycda@kaist.ac.kr}\\
$^{\dagger}$~Corresponding authors.}

\maketitle
\vspace{-5mm}

\begin{abstract}
Existing person search methods assume access to complete visual queries or 
exhaustive tracking, yet real-world witness accounts are vague, partial, 
and spread across cameras and time. 
We introduce ARGOS (Agentic Retrieval with Grounded Observational Search)\blfootnote{Code and data: \url{https://github.com/gritYCDA/ARGOS}}, 
a benchmark and agent framework that recasts multi-camera person search from 
one-shot retrieval on a complete query into interactive reasoning from partial 
clues. To our knowledge, ARGOS is the first interactive benchmark to couple 
witness dialogue with camera-network topology, requiring an agent to plan, 
question, and eliminate under information asymmetry.

An ARGOS agent receives a vague witness statement and must decide what to ask, 
when to invoke spatial or temporal tools, and how to interpret ambiguous 
natural-language responses, all within a limited turn budget.
To ground reasoning in physical constraints, the agent accesses a 
Spatio-Temporal Topology Graph (STTG) encoding camera connectivity and 
empirically validated transition times.
The benchmark comprises 2{,}691 tasks across 14 real-world scenarios in three 
progressive tracks: semantic perception (\emph{Who}, 989 tasks), spatial reasoning 
(\emph{Where}, 550 tasks), and temporal reasoning (\emph{When}, 1{,}152 tasks).
We propose Turn-Weighted Success (TWS) as the primary metric, jointly measuring 
correctness and turn efficiency.
Experiments with four LLM backbones show the benchmark is far from solved: 
the best agent achieves TWS of 0.383 (Track~2) and 0.590 (Track~3).
Ablations confirm each component is essential: removing domain-specific tools 
drops Top-1 accuracy by up to 49.6 percentage points, and removing strategic 
reasoning halves TWS while barely affecting Top-1.

\end{abstract}

\vspace{-5mm}
\section{Introduction}
\vspace{-2mm}
\label{sec:intro}Identifying a target person of interest in a large physical area is a fundamental requirement in intelligent surveillance systems, such as tracking a criminal suspect or locating a missing child in complex environments like large buildings or urban spaces. Consequently, this problem has been extensively studied in the computer vision community.

\begin{figure*}[t]
  \centering
  \includegraphics[width=\textwidth]{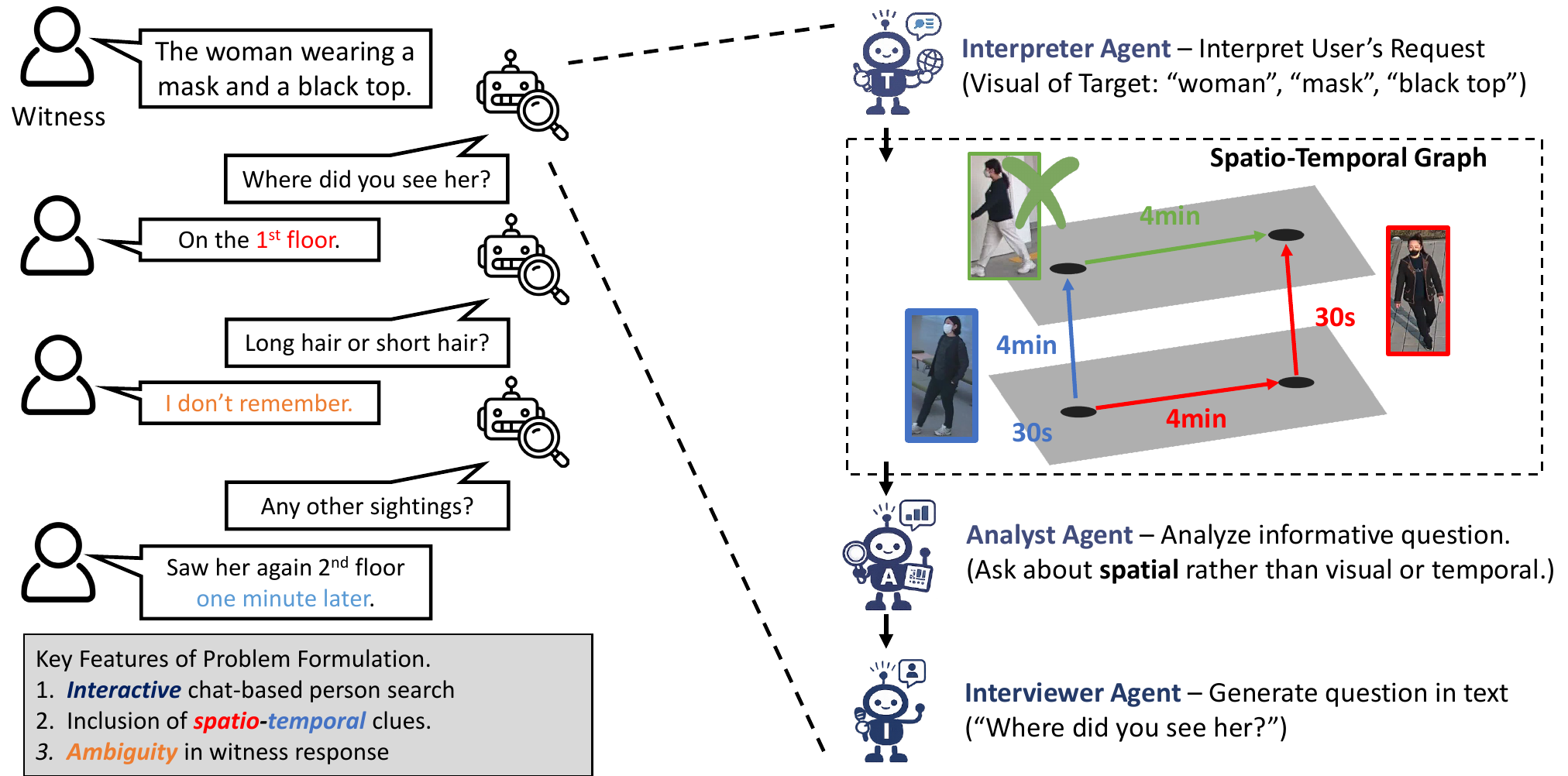}
  \caption{\textbf{Interactive multi-camera person search.}
  Given only a vague witness statement, the ARGOS agent asks targeted
  questions and issues spatial and temporal tool calls over multiple turns
  to locate the target across a 16-camera network.
  This couples interactive dialogue with spatio-temporal reasoning,
  which prior person-search formulations handle separately.}
  \label{fig:teaser}
  \vspace{-6mm}
\end{figure*}

In the early stage, the traditional setup of person re-identification (ReID)~\cite{zheng2015scalable,ye2021deep,wei2018person} aims to identify a target person from a gallery constructed from multiple CCTV cameras deployed across different locations. In this setting, the system searches for the same individual based on direct visual observations captured at a specific time and location, typically provided as an image crop of the target person.
Despite the tremendous advances, such approaches still exhibit a fundamental gap from real-world scenarios, as they rely on clear visual observations of the target person, which are often unavailable in urgent situations.

To bridge the gap between academic settings and real-world needs, several new formulations of the target search problem have recently emerged. One notable direction is text-based person ReID~\cite{yang2023towards,tan2024harnessing}, which relaxes the requirement of visual observations and instead assumes that witnesses provide textual descriptions of the target person's visual appearance. Building upon this direction, interactive person retrieval~\cite{lu2025llava,bai2025chat} has been proposed, in which a witness provides partial and potentially vague descriptions through multi-turn interactions. The system iteratively refines the candidate pool based on the ongoing dialogue.

Despite these advances toward more realistic formulations, existing approaches remain limited along two critical dimensions.
First, they operate on \emph{static queries}, with no mechanism for an agent to actively plan which questions to ask.
Second, they rely solely on visual appearance, neglecting spatial and temporal cues that are central to real-world investigations: witnesses routinely report \emph{where} and \emph{when} a person was last seen, and physical constraints on movement between cameras can rule out candidates entirely.
These two gaps define an unoccupied position at the intersection of interactive dialogue and multi-camera spatio-temporal reasoning.

In this paper, we introduce \textbf{ARGOS} (Agentic Retrieval with Grounded Observational Search), a benchmark and agent framework that reformulates person search as an active reasoning problem under information asymmetry (Figure~\ref{fig:teaser}). In ARGOS, an agent receives a vague initial witness statement about a target identity and must decide what to ask, when to use spatial or temporal tools, and how to interpret ambiguous responses, all within a limited turn budget. The agent is grounded by a Spatio-Temporal Topology Graph (STTG) that encodes physically validated camera-to-camera transition statistics, enabling it to eliminate candidates whose movements are infeasible.

Our contributions are as follows:

\vspace{-3mm}

\begin{itemize}
\item We define \textbf{interactive multi-camera person search}, a new task at the intersection of multi-turn witness interaction and spatio-temporal reasoning over camera networks. While spatio-temporal filtering is long established in person re-identification, ARGOS is the first to place it under interactive agent control for multi-camera person search, where the agent decides what to ask, when to test physical feasibility, and how to interpret ambiguous responses.

\item We construct the \textbf{ARGOS benchmark}, comprising 2{,}691 tasks across 14 real-world scenarios in three progressive tracks: semantic perception (\emph{Who}, 989 tasks), spatial reasoning (\emph{Where}, 550 tasks), and temporal reasoning (\emph{When}, 1{,}152 tasks) over directly observed camera pairs. Each track isolates a distinct capability while sharing a common gallery, evaluation protocol, and agentic metrics.

\item We introduce the \textbf{Spatio-Temporal Topology Graph (STTG)}, a directed weighted graph built from multi-camera tracking data that encodes camera connectivity and transition-time statistics. The STTG serves as both the structural backbone for task construction and the agent's grounding tool for temporal feasibility reasoning.

\item We provide a \textbf{strong baseline agent} equipped with a four-module pipeline and eight environment tools, demonstrating that tool-augmented agentic interaction substantially outperforms direct LLM inference for person search. Experiments across four LLM backbones confirm the benchmark is far from solved (best TWS: 0.383 on Track~2, 0.590 on Track~3) and that removing tools or strategic planning causes performance to collapse.
\end{itemize}

\section{Related work}
\vspace{-2mm}
\subsection{Language-based Person Re-identification}
\vspace{-2mm}

Language-based person re-identification aims to identify a target person based on natural language descriptions.
It can generally be categorized into two groups: (1) text-to-image Re-ID~\cite{li2017person,yang2023towards,tan2024harnessing,cao2024empirical,niu2025chatreid} and (2) interactive Re-ID~\cite{lu2025llava,bai2025chat}. These two setups mainly differ in whether the interaction between the system and the user occurs in a single turn or multiple turns. In text-to-image Re-ID, a single textual query is assumed to contain sufficient information about the target's visual appearance. Consequently, most methods focus on constructing robust vision--language joint embeddings that align images with their corresponding textual descriptions, building on CLIP~\cite{radford2021learning} and progressing from global~\cite{farooq2020convolutional,zheng2020dual} to fine-grained, multi-granularity alignment~\cite{chen2022tipcb,gao2022conditional,shao2022learning,fujii2023bilma,jiang2023cross}.

In contrast, interactive Re-ID simulates multi-round interactions between the system and the witness~\cite{das2017visual}. The system generates questions about the target person, and the witness provides answers that progressively refine the description. These approaches typically leverage multimodal large language models~\cite{grattafiori2024llama,chen2024internvl,li2024llava} to handle multi-turn dialogues and generate informative questions that narrow down the candidate pool.

However, these methods rely solely on visual appearance, neglecting where and when the target was observed. We instead propose \emph{multi-turn, spatio-temporal aware interactive person search}, an agentic setting in which the agent fuses visual, spatial, and temporal evidence across a camera network through dialogue rather than a single static query.

\vspace{-3mm}
\subsection{Spatial-Temporal Person Re-identification}
\vspace{-2mm}

A parallel line of work constrains person retrieval with camera topology and transition times. Wang~\etal~\cite{wang2019spatial} estimate a camera-to-camera transition-time distribution that discounts temporally infeasible gallery matches, and Cho~\etal~\cite{cho2019joint} jointly infer camera-network topology and cross-view identities. In both, a single transition distribution is fixed at training time and applied as a post-hoc filter over a static gallery. ARGOS moves this decision to inference time: the agent chooses when to test temporal feasibility and which sightings to query, turning the same physical constraint from a retrieval post-process into an interactive planning step.

\vspace{-3mm}
\subsection{Interactive Cross-Modal Retrieval}
\vspace{-2mm}

Interactive cross-modal retrieval iteratively searches for a target through multi-turn user interactions, where the user provides feedback~\cite{rui1998relevance,wu2004willhunter} or clarifications~\cite{cai2021ask,das2017learning,madasu2022learning}. Recent works leverage LLMs~\cite{achiam2023gpt} for interactive retrieval of images~\cite{levy2023chatting,lee2024interactive,zhu2024enhancing} and videos~\cite{han2024merlin,han2025ivcr,zhang2025quantifying,liang2023simple}.
ChatIR~\cite{levy2023chatting} introduced a chat-based image retrieval protocol where an LLM refines the user's intent through dialogue. PlugIR~\cite{lee2024interactive} generates questions that target current retrieval bottlenecks. In video retrieval, UMIVR~\cite{zhang2025quantifying} models multiple sources of uncertainty to improve question quality.

Unlike these systems, which refine a representation to improve ranking, ARGOS treats dialogue as an investigative process: the agent collects evidence, invokes spatio-temporal tools, and eliminates candidates through grounded reasoning rather than embedding refinement.

\vspace{-3mm}
\subsection{Agentic Video Understanding}
\vspace{-2mm}

Recent work equips LLM-based agents with tools to reason over long-form video.
VideoAgent~\cite{wang2024videoagent,fan2024videoagent} formulates video understanding as iterative frame retrieval guided by an LLM planner.
EGAgent~\cite{rege2025egagent} introduces entity scene graphs as structured memory for multi-hop reasoning over very long egocentric streams, and DVD~\cite{chu2025dvd} further scales agentic search to hour-long videos through multi-granular tool use.

These systems share our tool-augmented iterative reasoning but target video QA on single streams, without multi-camera topology or transition constraints. ARGOS instead identifies persons across a camera network, grounded by the STTG, and adds a benchmark whose metric (TWS) jointly measures correctness and interaction efficiency.

\vspace{-2mm}
\section{The ARGOS Benchmark}
\label{sec:argos}
\vspace{-2mm}
\subsection{Problem Formulation}
\label{sec:formulation}
\vspace{-2mm}

Standard text-based person retrieval takes a fixed natural-language description~$q$
and ranks a gallery $\mathcal{G}=\{g_1,\dots,g_N\}$ by visual similarity.
The query is given once, with no mechanism for the system to request clarification.
This formulation assumes the description suffices to identify the target---an
assumption that rarely holds in surveillance, where witnesses provide partial,
ambiguous, and incrementally recalled information across multiple cameras and time
periods.

We formulate \textbf{spatio-temporal aware interactive multi-camera person retrieval}
as follows.
An agent~$\pi$ receives an initial statement from a witness about a target
$g^*\!\in\mathcal{G}$ and engages in a multi-turn dialogue with a witness
simulator~$\mathcal{W}$.
At each turn~$t$ the agent selects an action~$a_t$---asking about visual attributes,
querying spatial location, or invoking temporal reasoning tools---and the witness
responds according to $g^*$'s ground-truth attributes and trajectory.
The agent also accesses a Spatio-Temporal Topology Graph
$\mathcal{T}$ that encodes camera connectivity and transition-time statistics.

Let $\mathcal{C}_0\subseteq\mathcal{G}$ denote the initial candidate set consistent
with the opening statement, and $\mathcal{C}_t$ the set after turn~$t$.
A task is solved when $\mathcal{C}_t=\{g^*\}$.
Performance is measured by correctness
($g^*\!\in\mathcal{C}_{\mathrm{final}}$, $|\mathcal{C}_{\mathrm{final}}|=1$)
and efficiency (turn count~$\tau$ relative to the oracle-optimal~$\tau^*$).

This formulation differs from prior interactive retrieval
work~\cite{bai2025chat,lu2025llava} in three respects.
First, the query arrives incrementally---the agent decides \emph{what to ask}.
Second, the gallery spans multiple cameras with known spatial topology, enabling
location-based reasoning.
Third, temporal constraints (transition times between cameras) allow the agent to
eliminate candidates whose movements are physically impossible.
Existing interactive methods~\cite{lu2025llava,bai2025chat,levy2023chatting,lee2024interactive}
build on single-view re-ID or image retrieval datasets that structurally lack
multi-camera spatial and temporal information; ARGOS builds on
MTMMC~\cite{woo2024mtmmc}, whose synchronized multi-camera video, tracking
annotations, and instance labels make such reasoning possible.

\vspace{-2mm}
\subsection{Benchmark Overview}
\label{sec:overview}
\vspace{-2mm}

ARGOS consists of 2{,}691 task instances across 14 scenarios, organized into three
tracks that test progressively complex reasoning (Table~\ref{tab:stats}).
All tasks draw from a common gallery of 1{,}273 annotated persons observed by
16 synchronized cameras in two environments: an industrial factory
(7~scenarios, summer) and a university campus (7~scenarios, autumn),
sourced from the MTMMC dataset~\cite{woo2024mtmmc}.
Each person is annotated with 24 visual attributes
(Sec.~C).
On average, each person is observed by 5.9 cameras (range 4.9--7.4),
providing rich multi-camera trajectories for spatial and temporal task construction.

\begin{table}[t]
\centering
\caption{ARGOS benchmark statistics across three tracks.}
\vspace{-3mm}
\label{tab:stats}
\resizebox{\linewidth}{!}{%
\begin{tabular}{lccc}
\toprule
& \textcolor{argosWho}{\textbf{Track\,1 (\emph{Who})}}
& \textcolor{argosWhere}{\textbf{Track\,2 (\emph{Where})}}
& \textcolor{argosWhen}{\textbf{Track\,3 (\emph{When})}} \\
\midrule
Task count              & 989 & 550 & 1{,}152 \\
Interaction mode        & Single-turn & Multi-turn & Multi-turn \\
Core capability         & Semantic parsing & Spatial reasoning & Temporal reasoning \\
Avg.\ dialogue context  & 5.6 turns & --- & --- \\
Avg.\ oracle turns\,($\tau^*$) & --- & 2.02 & 1.89 \\
Primary metric          & Top-1 Acc & TWS & TWS \\
\bottomrule
\end{tabular}}
\vspace{-5mm}
\end{table}

\vspace{-3mm}
\paragraph{Dataset construction.}
All three tracks share a two-stage pipeline
(Sec.~A).
Stage~1 generates deterministic ground truth through algorithmic computation:
information-theoretic clue selection for Track~1, zone-based spatial disambiguation
for Track~2, and STTG-based temporal feasibility classification for Track~3.
No language model is involved; task validity and optimal turn sequences are computed
from the attribute database, camera topology, and trajectory records.
Stage~2 wraps the structured ground truth in natural-language dialogues via an LLM,
converting attribute labels and camera identifiers into conversational witness
statements.
This separation ensures benchmark correctness does not depend on language-model
behavior, while the natural-language surface prevents agents from bypassing
comprehension by directly matching structured labels.

\subsection{Spatio-Temporal Topology Graph}
\label{sec:sttg}
\begin{figure}[t]
  \centering
  \includegraphics[width=\textwidth]{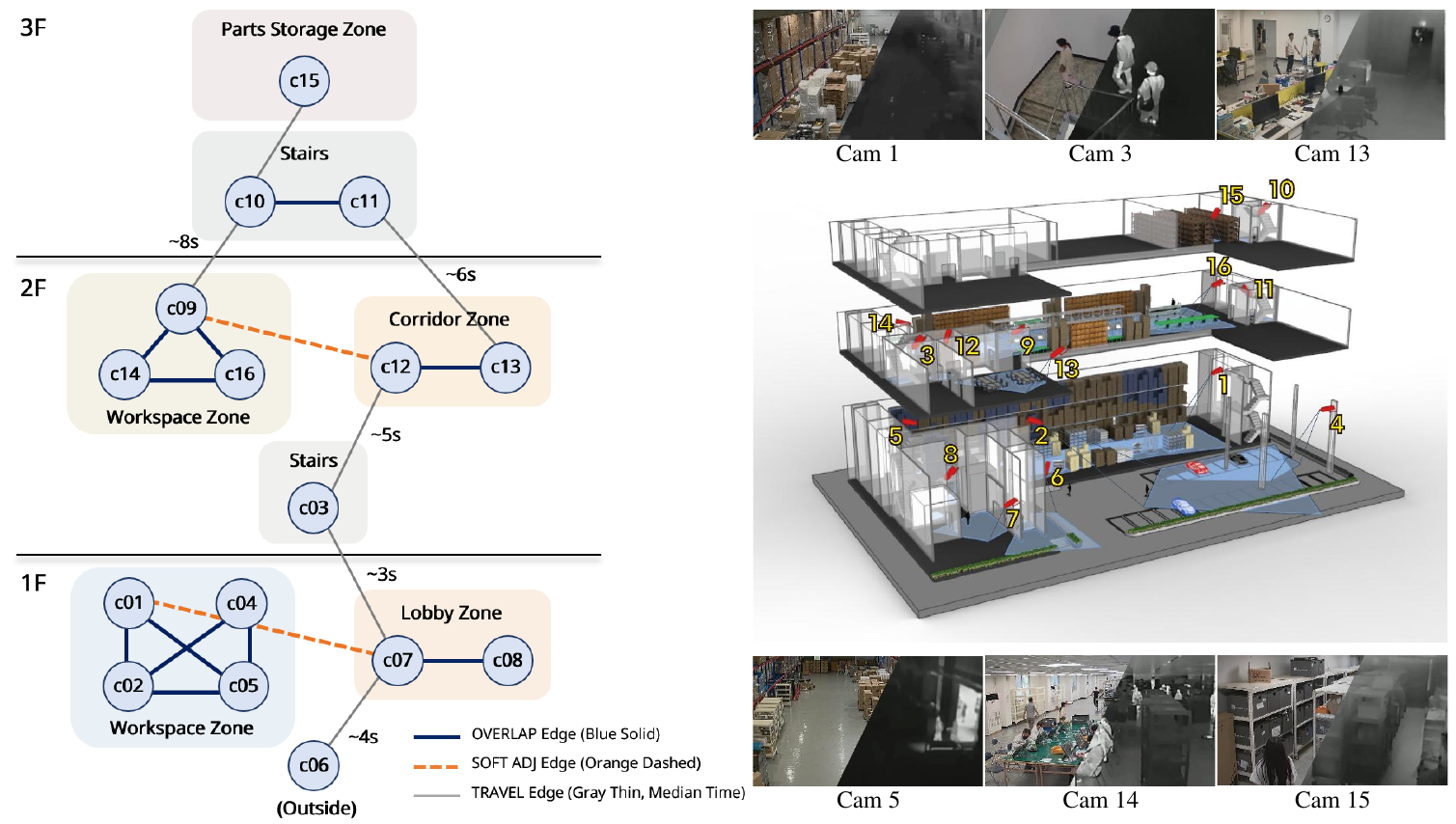}
  \vspace{-7mm}
  \caption{\textbf{Left:} STTG for the factory (16 cameras).
   Nodes are grouped into zones (shaded) by \texttt{OVERLAP} connectivity.
   Edge types: \texttt{OVERLAP} (blue), \texttt{SOFT\_ADJ} (orange),
   \texttt{TRAVEL} (gray); labels show median transition time.
   \textbf{Right:} 3-D camera layout with sample RGB--Thermal pairs.
   The university STTG (Sec.~C) has 149 edges.}
  \label{fig:sttg}
  \vspace{-5mm}
\end{figure}

The STTG encodes camera relationships in space and time.
It serves two roles: the benchmark derives ground-truth tasks from it
(Tracks~2 and~3), and agents receive it as an environment representation for
spatio-temporal reasoning.

\vspace{-2mm}
\paragraph{Definition.}
The STTG is a directed weighted graph $\mathcal{T}=(V,E)$ where each node
$v_i\in V$ represents a camera, annotated with a zone label and a natural-language
sub-area description (e.g., ``deep inside the warehouse, near the tall shelves'').
Each directed edge
$e_{ij}=(v_i,v_j,\lambda_{ij},\sigma_{ij})$
carries a type
$\lambda_{ij}\in\{\texttt{OVERLAP}, \allowbreak \texttt{SOFT\_ADJ},\texttt{TRAVEL}\}$
and transition-time statistics
$\sigma_{ij}=(t_{\min},t_{\mathrm{med}},t_{\max},n)$
computed from observed movements.
\texttt{OVERLAP} edges connect cameras with shared fields of view (near-zero
transition time); their connected components define \emph{zones}.
\texttt{SOFT\_ADJ} edges connect cameras separated by a door or passage with no
visual overlap but small transit ($<\!2$\,s).
\texttt{TRAVEL} edges cover all remaining transitions, with times ranging from
seconds to minutes.

\vspace{-2mm}
\paragraph{Construction.}
The STTG is built from multi-camera tracking ground truth through a three-phase
pipeline: transition extraction, priority-based labeling with human verification, and
directed-edge aggregation
(Sec.~A).
The resulting graphs contain 110 edges (factory) and 149 edges (university),
all satisfying $t_{\min}\!\leq t_{\mathrm{med}}\!\leq t_{\max}$
(Fig.~\ref{fig:sttg}).

\vspace{-3mm}
\paragraph{Zones.}
\texttt{OVERLAP} components define atomic zones (e.g., a 4-camera warehouse zone,
a 6-camera building-core zone).
\texttt{SOFT\_ADJ} edges link adjacent zones into composite zones corresponding to
floors or building sections.
This bottom-up derivation grounds zone membership in measured camera co-visibility
rather than architectural assumptions.

\subsection{Three Tracks}
\label{sec:tracks}

ARGOS defines three tracks that isolate progressively complex capabilities
(Fig.~\ref{tab:stats}).
All tracks share the same gallery, attribute schema, and scenarios.
Track~1 can be evaluated independently; Tracks~2 and~3 presuppose Track~1's
semantic parsing.


\vspace{-2mm}
\paragraph{Track~1: \emph{Who}---Semantic Perception (989 tasks).}
The agent receives a completed multi-turn dialogue
(avg.\ 5.6 exchanges) in which a witness describes the target's appearance.
The dialogue is generated by converting an optimal attribute-elimination sequence
into natural language
(Sec.~A)
and contains synonyms, hedged responses, and implicit references that require robust
parsing.
The agent extracts attributes, filters the gallery, and returns a single prediction.
Each task targets a unique person; Track~1 serves as a prerequisite for Tracks~2
and~3.

\vspace{-2mm}
\paragraph{Track~2: \emph{Where}---Spatial Reasoning (550 tasks).}
The witness reports seeing the target in a multi-camera zone
(e.g., ``the warehouse area,'' spanning 4~cameras).
The agent resolves \emph{which sub-area} within the zone through
\textbf{ask-spatial} questions (e.g., ``Was it indoors or outdoors?''),
whose answers map to specific cameras via the zone's disambiguation
structure (Sec.~A),
and further narrows candidates via \textbf{ask-attribute} questions before
issuing a \textbf{predict} action.
Every task requires at least one spatial turn by construction.
The oracle path averages 2.02 turns (1.14 spatial, 0.91 attribute).

\vspace{-2mm}
\paragraph{Track~3: \emph{When}---Temporal Reasoning (1{,}152 tasks).}
The witness reports two sightings at different locations and approximate times
(e.g., ``in the warehouse right at the beginning, then on the second floor a few
minutes later'').
The agent applies a temporal elimination tool that evaluates each candidate against
the STTG: a candidate must have been present at both cameras and have a transition
time within the STTG's observed range.
Candidates are eliminated by \emph{time reversal} (arrived at the second camera
before leaving the first) or as \emph{too slow} (transition exceeds the observed
maximum).
Absence-based filtering (never observed at the reported camera) is applied as
preprocessing; Track~3 tasks specifically require at least one temporal elimination
beyond simple presence checks.
The oracle path averages 1.89 turns.

Only camera pairs with direct observed edges in the STTG generate temporal claims;
multi-hop route inference is excluded, ensuring every elimination rests on empirical
data.

\subsection{Evaluation Protocol}
\label{sec:eval}
\paragraph{Witness Simulator.}
\label{sec:eval_witness_simulator}
ARGOS evaluation uses a deterministic witness simulator~$\mathcal{W}$.
For the three observable attributes---visual gender, upper clothing color, and lower
clothing color, selected for high class variation and perceptual
salience~\cite{meissner2007person_descriptions}---$\mathcal{W}$ returns the
ground-truth value in varied natural-language templates
(Sec.~C).
For other attributes it responds with uncertainty; for spatial and temporal queries
it returns pre-computed responses from the ground-truth path.

This controlled-fidelity design ensures \textbf{reproducibility} (identical witness
behavior across runs) and \textbf{diagnostic clarity} (agent failures trace to
parsing, reasoning, or strategy---not witness noise).
The templates nonetheless require robust language understanding; an agent that cannot
parse hedged phrasing loses information at every turn.

\vspace{-3mm}
\paragraph{Metrics.}
For Track~1 (single-turn) the primary metric is \textbf{Top-1 Accuracy}.
An auxiliary \textbf{Parsing Accuracy} measures attribute extraction quality
independent of retrieval success.
For Tracks~2 and~3 (multi-turn) the primary metric is
\textbf{Turn-Weighted Success} (TWS):
\begin{equation}\label{eq:tws}
  \text{TWS}
  = \frac{1}{N}\sum_{i=1}^{N} s_i \cdot
    \frac{\tau_i^*}{\max(\tau_i,\,\tau_i^*)}
\end{equation}
where $s_i\!\in\!\{0,1\}$ indicates correctness, $\tau_i$ is the agent's turn count,
and $\tau_i^*$ is the oracle-optimal turn count pre-computed for every task.
An agent solving every task at exactly the oracle count achieves $\text{TWS}=1.0$.
TWS follows the design principle of SPL~\cite{anderson2018evaluation} in embodied
navigation---weighting success by efficiency---but replaces path length with
dialogue turns.
Because $\tau^*$ is embedded in the dataset, evaluation requires no oracle
computation at test time.
TWS decomposes into \textbf{Top-1 Accuracy} (correctness) and
\textbf{Average Turns} on successful tasks (efficiency), both reported alongside TWS.
Additional diagnostic metrics (\textbf{SR\_turn@$T$}, per-scenario breakdowns) are
provided in Sec.~B.

\vspace{-2mm}
\section{ARGOS Agent: A Strong Baseline}
\label{sec:agent}
\vspace{-2mm}
To establish a reference point for the benchmark, we introduce ARGOS Agent, a tool-augmented LLM agent that interacts with the evaluation environment through a structured observe-think-act loop. Unlike existing person search systems that operate in a single forward pass, ARGOS Agent couples multi-turn investigative dialogue with tool-grounded spatio-temporal reasoning. The agent is designed to be strong yet transparent: it achieves competitive performance across all three tracks while exposing clear bottlenecks for future work.

\vspace{-4mm}
\subsection{System Architecture}
\label{sec:architecture}
\begin{figure}[t]
\centering
\includegraphics[width=0.9\linewidth]{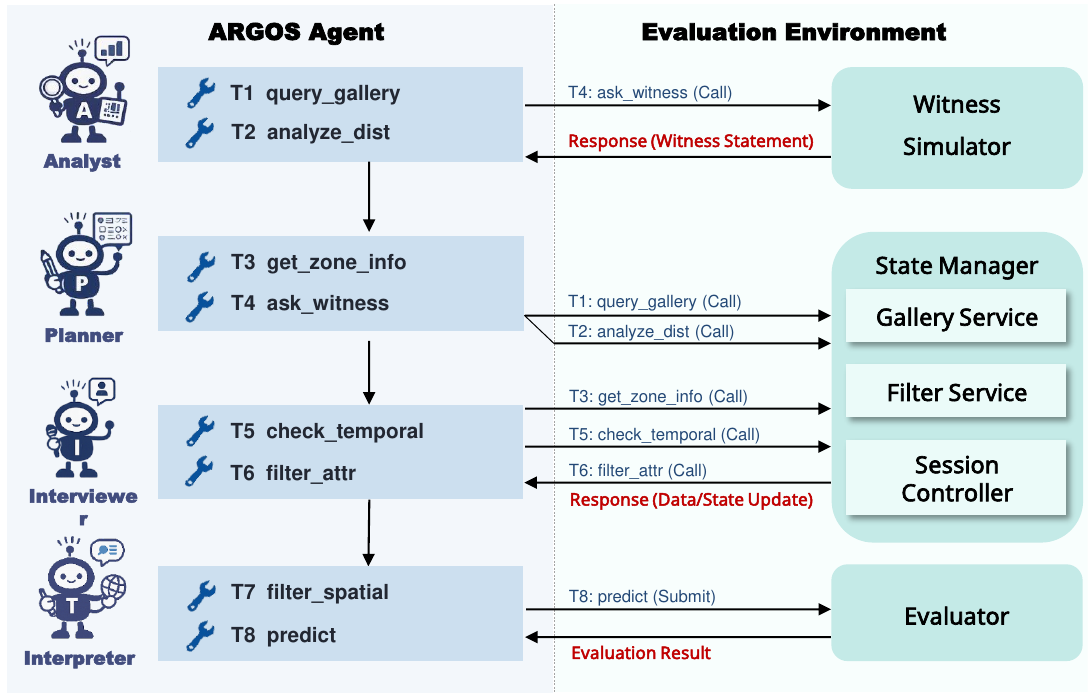}
\caption{System architecture. \textbf{Left}: the ARGOS Agent consists of four LLM-driven modules that form an observe-think-act loop. \textbf{Right}: the evaluation environment provides a Witness Simulator (Sec.~\ref{sec:eval_witness_simulator}) and a deterministic State Manager. Arrows denote tool calls (downward) and responses (upward). The agent selects actions; the environment executes them.}
\label{fig:architecture}
  \vspace{-6mm}
\end{figure}

Figure~\ref{fig:architecture} shows the separation between the \textbf{agent} (LLM-driven) and the \textbf{environment} (deterministic).

\vspace{-3mm}
\paragraph{Environment.}
The environment consists of two components.
The \emph{Witness Simulator}, defined in Sec.~\ref{sec:eval_witness_simulator}, returns natural-language answers to the agent's questions.
The \emph{State Manager} provides three deterministic services: a Gallery Service that exposes the attribute database (read-only), a Filter Service that performs exact-match attribute filtering and spatial camera mapping, and a Session Controller that enforces the 20-turn budget and termination conditions.
All State Manager responses are deterministic and require no LLM reasoning; the agent selects actions and the environment executes them, following a standard \texttt{env.step(action)} paradigm.

\vspace{-2mm}
\paragraph{Agent.}
The agent processes each turn through four modules in sequence:
\vspace{-1mm}

\begin{enumerate}
\item \textbf{Analyst} queries the gallery, computes attribute distributions over the current candidate set, and retrieves zone structure (Track~2). It identifies which attributes have the highest elimination power among remaining candidates.
\item \textbf{Planner} receives the Analyst's summary together with the full dialogue history and decides the next action: ask about an attribute, request a spatial description, or issue a temporal check.
\item \textbf{Interviewer} executes the chosen action by invoking the appropriate tool. For Track~3, \texttt{check\_temporal} (T5) is available in the Planner's action menu from the first turn, alongside attribute and spatial queries, and the agent falls back to attribute questions when a temporal cue cannot be parsed. We do not fix a call order: the ablation in Sec.~\ref{sec:ablation} shows that letting the agent decide when to invoke \texttt{check\_temporal} performs on par with or slightly better than invoking it early (Table~\ref{tab:autonomous}), so the Track~3 improvement reflects the tool's informational value rather than a fixed scaffold.

\item \textbf{Interpreter} parses the witness's natural-language response into a canonical attribute value and applies the corresponding filter to update the candidate set.
\end{enumerate}

This sequential pipeline runs until the agent issues a prediction or exhausts the turn budget. A critical design property is that the same tool set serves all three tracks, yet the Planner must make qualitatively different strategic decisions in each: in Track~1, the agent extracts attributes from a single dialogue turn; in Track~2, it balances attribute questions against a one-time spatial query; in Track~3, it must integrate STTG-based temporal elimination with attribute-based narrowing. Fig~\ref{tab:info_boundary} makes the agent's information boundary explicit.

\begin{table}[t]
\centering
\caption{\textbf{Information boundary of the ARGOS Agent.} The agent cannot observe
which attributes the witness will answer, forcing strategic decisions under
uncertainty.}
\vspace{-3mm}
\label{tab:info_boundary}
\small
\setlength{\tabcolsep}{6pt}
\renewcommand{\arraystretch}{1.15}
\begin{tabularx}{\linewidth}{@{}>{\raggedright\arraybackslash}X>{\raggedright\arraybackslash}X@{}}
\toprule
\textbf{Accessible} & \textbf{Not accessible} \\
\midrule
Gallery DB (24 attributes, read-only) & Candidate--camera mapping \\
Current candidate ID list & Ground-truth disambiguation path \\
Zone structure (sub-areas, cameras) & Which attributes the witness can answer \\
STTG statistics (min, median, max, std) & Per-candidate feasibility details \\
Available action types & Witness's internal observable set \\
\bottomrule
\end{tabularx}
\end{table}


The last row is particularly consequential. The agent does not know that the witness can only report on three observable attributes (gender, upper-body color, lower-body color) out of 24 in the gallery; if the highest-elimination-power attribute happens to be non-observable, the turn yields no information, forcing the agent to adapt its strategy from failed queries.

\subsection{Tool Definitions}
\label{sec:tools}

Table~\ref{tab:tools} lists the eight tools available to the agent. Tools T1--T3 support analysis, T4--T5 support interaction with the environment, and T6--T8 produce filtering or prediction actions.

\begin{table}[t]
\centering
\caption{\textbf{Tool registry.} The agent invokes eight tools via structured
function calls, each executed by the environment (State~Manager or Witness
Simulator). Tool names are tinted by track role
(\textcolor{argosWho}{\textbf{Who/attribute}},
\textcolor{argosWhere}{\textbf{Where/spatial}},
\textcolor{argosWhen}{\textbf{When/temporal}}; shared tools stay neutral).
Parsing the witness reply at \texttt{filter\_attr} is the primary bottleneck:
free-form answers must map to an exact canonical value, and errors propagate into
filtering.}
\vspace{-3mm}
\label{tab:tools}
\footnotesize
\setlength{\tabcolsep}{4pt}
\renewcommand{\arraystretch}{1.1}
\begin{tabularx}{\linewidth}{@{}c l l >{\raggedright\arraybackslash}X c@{}}
\toprule
\textbf{ID} & \textbf{Name} & \textbf{Input} & \textbf{Output} & \textbf{Tracks} \\
\midrule
T1 & \texttt{query\_gallery} & candidate IDs & attribute records & 1,2,3 \\
T2 & \texttt{analyze\_dist} & candidate IDs & counts, elim.\ power & 1,2,3 \\
T3 & \texttt{\textcolor{argosWhere}{get\_zone\_info}} & --- & sub-areas, camera list & 2 \\
T4 & \texttt{ask\_witness} & question type & NL response (string) & 2,3 \\
T5 & \texttt{\textcolor{argosWhen}{check\_temporal}} & candidate IDs & filtered partition & 3 \\
T6 & \texttt{\textcolor{argosWho}{filter\_attr}} & attr, value, IDs & filtered IDs & 1,2,3 \\
T7 & \texttt{\textcolor{argosWhere}{filter\_spatial}} & camera ID, IDs & filtered IDs & 2 \\
T8 & \texttt{predict} & candidate ID & correctness flag & 1,2,3 \\
\bottomrule
\end{tabularx}
\end{table}


Two aspects of this tool design merit attention.
First, T5 (\texttt{check\_temporal}) provides the agent's interface to the Spatio-Temporal Topology Graph introduced in Sec.~\ref{sec:sttg}. Given the current candidate set and a reference timestamp, the State Manager queries the STTG and returns the subset of candidates whose traversal times are feasible. This single call can eliminate a substantial fraction of candidates without any dialogue, making it the most information-dense tool in the registry.
Second, T6 (\texttt{filter\_attr}) takes as input a canonical attribute value that the agent must extract from the witness's natural-language response. Because the witness uses varied phrasings (12 response templates and, in Track~2, GPT-generated conversational responses), natural-language parsing is a non-trivial step where errors directly degrade the candidate set. We quantify this bottleneck in Sec.~\ref{sec:experiments}.

\vspace{-2mm}
\section{Experiments}\label{sec:experiments}

\subsection{Experimental Setup}\label{sec:exp_setup}

We evaluate across all three tracks using four LLM backbones
(GPT-5.2, GPT-4o, GPT-5-mini, Claude~Sonnet~4)
at temperature~0.0 with a maximum of 20 turns per episode and seed~42.
TWS (Sec.~\ref{sec:eval}), which jointly measures accuracy and turn efficiency,
serves as the primary metric for Tracks~2 and~3;
Top-1 accuracy and SR@$K$ are reported for Track~1.
Ablation baselines are specified in Table~\ref{tab:ablation},
where each row names the component removed from the full agent.

\subsection{Main Results}\label{sec:main_results}

\paragraph{Track~1.}
Structured tool-calling outperforms end-to-end LLM inference (Table~\ref{tab:task1}).
LLM~ToolCall achieves 81.1\% SR@1, +7.8\,pp over LLM~Direct (73.3\%).
Although LLM~Direct attains slightly higher parsing accuracy (93.0\% vs.\ 90.8\%),
it cannot recover from filtering errors and produces only a single unranked prediction.

\begin{table}[t]
\centering
\caption{%
\textbf{Track~1 results} (GPT-4o).
LLM~Direct outputs a single ID (SR@1\,=\,SR@5).
}
\vspace{-3mm}
\label{tab:task1}
\small
\begin{tabular}{lccc}
\toprule
\textbf{Agent} & \textbf{SR@1} & \textbf{SR@5} & \textbf{Parse Acc.} \\
\midrule
Oracle          & 100.0 & 100.0 & 100.0 \\
LLM ToolCall    &  81.1 &  82.3 &  90.8 \\
LLM Direct      &  73.3 &  73.3 &  93.0 \\
Rule-based      &  32.2 &  48.0 &  66.0 \\
\bottomrule
\end{tabular}
\\[1pt]
{\footnotesize (\%) SR@$K$: success rate within top-$K$.}
\end{table}

\paragraph{Tracks~2 and 3.}

The benchmark remains far from solved:
the best TWS reaches 0.383 on Track~2 (Claude~Sonnet~4) and 0.590 on Track~3 (GPT-5.2),
both well below the Oracle (Table~\ref{tab:main}).
No single backbone dominates both tracks:
Claude~Sonnet~4 leads Track~2 but ranks last on Track~3, while GPT-5.2 shows the reverse.
This crossover indicates that spatial and temporal reasoning stress different capabilities.
The SR$_t$@5 column reveals that Claude~Sonnet~4 resolves Track~2 tasks earliest,
consistent with its TWS lead.

\begin{table}[t]
\centering
\caption{%
\textbf{Main results on Tracks~2 and 3} across four LLM backbones.
Best TWS per track is \textbf{bolded}.
\colorbox{blue!8}{Blue shading} highlights the best value in each column.
}
\vspace{-3mm}
\label{tab:main}
\small
\setlength{\tabcolsep}{3.5pt}
\begin{tabular}{lcccccccc}
\toprule
& \multicolumn{4}{c}{\textbf{Track 2 (Spatial)}}
& \multicolumn{4}{c}{\textbf{Track 3 (Temporal)}} \\
\cmidrule(lr){2-5} \cmidrule(lr){6-9}
\textbf{Backbone}
  & TWS & Top-1 & SR$_t$@5 & AvgT$_s$
  & TWS & Top-1 & SR$_t$@5 & AvgT$_s$ \\
\midrule
Oracle
  & 1.000 & 100.0 & 100.0 & 2.05
  & 1.000 & 100.0 & 100.0 & 1.88 \\
\midrule
GPT-5.2
  & 0.338 & 73.1 & 33.6 & 7.04
  & \cellcolor{blue!8}\textbf{0.590} & \cellcolor{blue!8}88.2 & \cellcolor{blue!8}65.8 & \cellcolor{blue!8}3.91 \\
GPT-4o
  & 0.323 & 74.5 & 31.6 & 7.49
  & 0.567 & 80.6 & 60.8 & 3.91 \\
GPT-5-mini
  & 0.319 & 74.9 & 32.2 & 7.48
  & 0.556 & 88.0 & 60.5 & 4.40 \\
Claude Sonnet 4
  & \cellcolor{blue!8}\textbf{0.383} & \cellcolor{blue!8}76.0 & \cellcolor{blue!8}38.4 & \cellcolor{blue!8}6.71
  & 0.548 & 83.6 & 59.5 & 4.25 \\
\bottomrule
\end{tabular}
\\[1pt]
{\footnotesize
SR$_t$@$T$: success rate (budget $T$); AvgT$_s$: avg.\ turns (success cases)
}
\vspace{-3mm}
\end{table}

\vspace{-3mm}
\subsection{Ablation Study}\label{sec:ablation}
\begin{figure}[t]
  \centering
  \includegraphics[width=\linewidth]{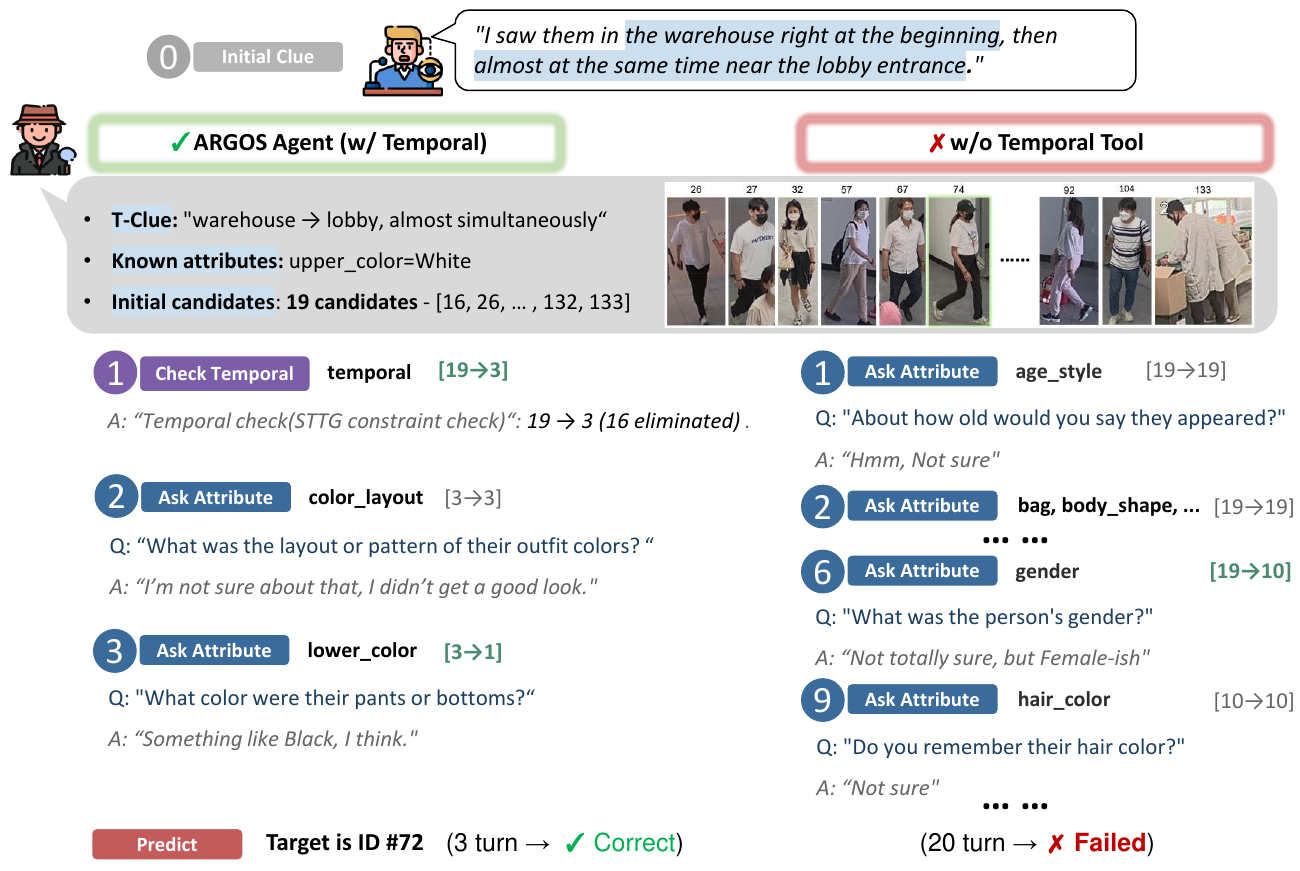}
  \vspace{-5mm}
  \caption{\textbf{Track~3 temporal-tool demonstration} (Factory, Hard).
  One \texttt{check\_temporal} call eliminates 16 of 19 candidates via STTG feasibility;
  the agent then identifies the target in 3 turns, while the ablated agent (no temporal tool)
  exhausts its 20-turn budget and fails.}
  \label{fig:temporal_demo}
\end{figure}

We ablate using GPT-4o (Table~\ref{tab:ablation}).
\vspace{-3mm}
\paragraph{(a) Domain-specific tools are indispensable.}
Without any tool (w/o All Tools), Track~3 Top-1 collapses to 11.4\%, on par with random guessing over the gallery.
Removing only the track-specific tool is equally destructive:
w/o Temporal Tool falls to 31.0\% ($-$49.6\,pp) and w/o Spatial Tool to 40.7\% ($-$33.8\,pp),
with TWS near zero in both cases.
The temporal drop is not an artifact of forcing this tool first: when \texttt{check\_temporal} is instead offered in the action menu for the agent to invoke at will, Track~3 Top-1 rises from 80.6\% to 83.9\% and TWS from 0.567 to 0.572 (Table~\ref{tab:autonomous}, Fig.~\ref{fig:temporal_demo}), so its contribution reflects the candidates it eliminates rather than its position in the sequence.

\begin{table}[t]
\centering
\caption{%
\textbf{Autonomous vs.\ temporal-first invocation} (GPT-4o).
Letting the agent decide when to call \texttt{check\_temporal} matches or slightly improves fixed early invocation, while removing the tool collapses Track~3.}
\vspace{-3mm}
\label{tab:autonomous}
\small
\renewcommand{\arraystretch}{0.9}
\setlength{\tabcolsep}{4pt}
\begin{tabular}{lccc}
\toprule
Track~3 variant & Top-1 & TWS & AvgTurns \\
\midrule
Temporal-first    & 80.6 & 0.567 & 3.91 \\
Autonomous        & \textbf{83.9} & \textbf{0.572} & 4.05 \\
w/o Temporal Tool & 31.0 & 0.054 & 13.44 \\
\bottomrule
\end{tabular}
\end{table}

\paragraph{(b) Strategic reasoning improves efficiency beyond accuracy.}
On Track~3, w/o Strategy nearly matches the full agent in Top-1 (76.9\% vs.\ 80.6\%),
yet its TWS is sharply lower (0.373 vs.\ 0.567) because it requires twice the turns (7.73 vs.\ 3.91).
Top-1 alone masks this efficiency gap; TWS captures it directly.
On Track~2, the accuracy gap itself is large (+26.9\,pp), and TWS compounds it further.

\paragraph{(c) Agentic interaction is the only viable path.}
On Track~3, both Single-Pass (Initial) and w/o All Tools converge near 11\% Top-1,
a structural floor replicated across all backbones.
Neither stronger LLMs nor richer prompts overcome this barrier; only the combination of tool access and multi-turn interaction reaches 80.6\%.
On Track~2, Single-Pass (Full) unexpectedly falls below Single-Pass (Initial)
(56.2\% vs.\ 64.7\%), a pattern consistent across two backbones,
which we attribute to parsing difficulty (Sec.~B).

\begin{table}[t]
\centering
\caption{%
\textbf{Ablation results} (GPT-4o).
Each row removes one component.
Single-pass baselines have no meaningful TWS or AvgT$_s$.
\colorbox{red!8}{Red shading} marks the steepest drop per track.
}
\vspace{-3mm}
\label{tab:ablation}
\small
\setlength{\tabcolsep}{3pt}
\begin{tabular}{llcccccc}
\toprule
& & \multicolumn{3}{c}{\textbf{Track 2}} & \multicolumn{3}{c}{\textbf{Track 3}} \\
\cmidrule(lr){3-5} \cmidrule(lr){6-8}
\textbf{Agent} & \textbf{Removed}
  & TWS & Top-1 & AvgT$_s$
  & TWS & Top-1 & AvgT$_s$ \\
\midrule
Oracle & \emph{(upper bound)}
  & 1.000 & 100.0 & 2.05
  & 1.000 & 100.0 & 1.88 \\
ARGOS Agent & \emph{(none)}
  & 0.323 & 74.5 & 7.49
  & 0.567 & 80.6 & 3.91 \\
\midrule
w/o Strategy & LLM reasoning
  & 0.136 & 47.6 & 10.78
  & 0.373 & 76.9 & 7.73 \\
w/o Spatial Tool & Spatial queries
  & \cellcolor{red!8}0.063 & \cellcolor{red!8}40.7 & 14.20
  & \multicolumn{3}{c}{---} \\
w/o Temporal Tool & Temporal queries
  & \multicolumn{3}{c}{---}
  & \cellcolor{red!8}0.054 & \cellcolor{red!8}31.0 & 13.44 \\
\midrule
Single-Pass (Full) & Interaction
  & --- & 56.2 & ---
  & --- & 29.6 & --- \\
Single-Pass (Initial) & Interaction + dialogue
  & --- & 64.7 & ---
  & --- & 11.3 & --- \\
w/o All Tools & All tools
  & \multicolumn{3}{c}{N/A}
  & --- & 11.4 & --- \\
\bottomrule
\end{tabular}
\vspace{-5mm}
\end{table}

\vspace{-3mm}
\subsection{Analysis}
\label{sec:analysis}

\paragraph{Failure patterns reveal distinct behavioral profiles.}
On Track~3, GPT-4o fails primarily through premature commitment:
91\% of its failures are wrong predictions, with negligible timeout (1.6\%).
Its turn count barely differs between successful and failed episodes (3.91 vs.\ 3.93 overall).
GPT-5.2 shows the opposite profile:
46\% of failures are timeouts from extended evidence gathering,
yielding fewer total errors and the highest TWS despite higher per-task latency.
These profiles suggest that TWS rewards sustained disambiguation
over rapid commitment.

\vspace{-2mm}
\paragraph{Tracks probe distinct capabilities.}
Track~2's bottleneck is natural-language understanding (78.6\% parse success vs.\ 94.2\% for Track~3; per-template breakdown in Sec.~B),
while Track~3's bottleneck is temporal reasoning (w/o Temporal Tool: $-$49.6\,pp, the steepest drop).
This explains the backbone crossover in Table~\ref{tab:main}
and confirms the two tracks evaluate complementary aspects of agentic person search.

\begin{figure}[t]
\centering
\includegraphics[width=\linewidth]{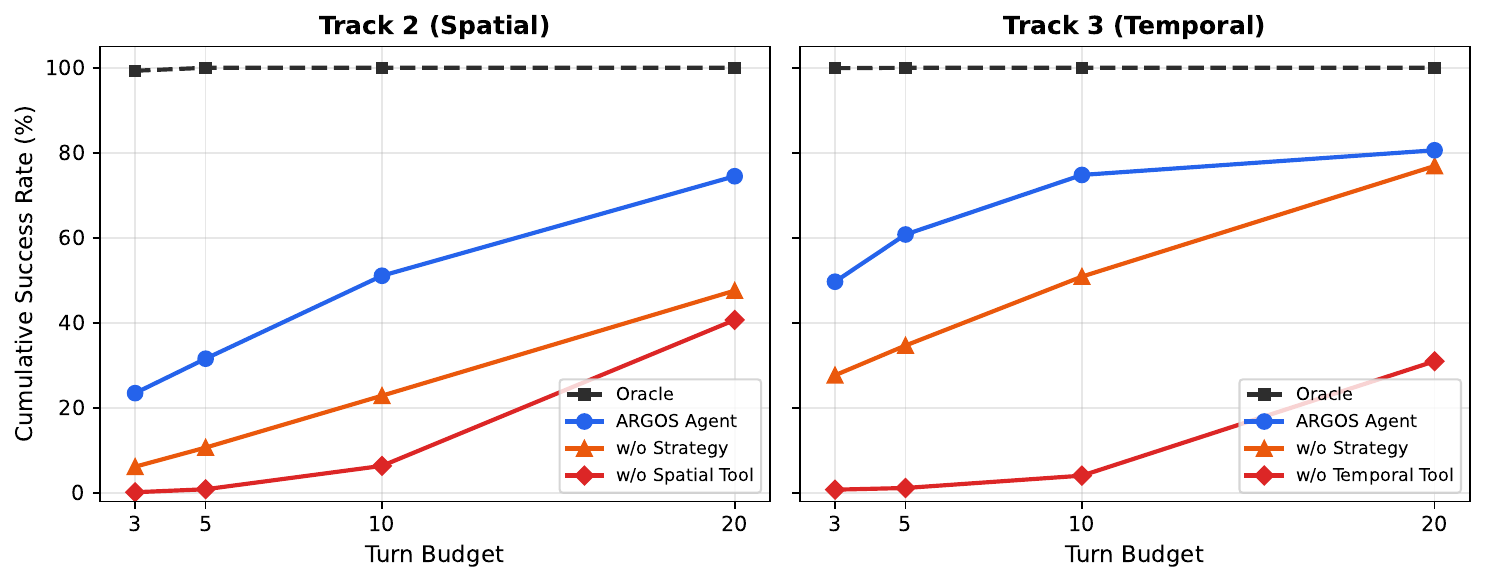}
\caption{%
\textbf{Cumulative success rate (SR$_t$@$T$) vs.\ turn budget} for Track~2 (left) and Track~3 (right), GPT-4o.
The ARGOS Agent (blue) reaches high accuracy faster than w/o Strategy (orange).
w/o Spatial/Temporal Tool (red) stays near zero through early turns.
}
\label{fig:sr_curve}
\vspace{-3mm}
\end{figure}

\vspace{-2mm}
\paragraph{Efficiency emerges from interaction.}
Figure~\ref{fig:sr_curve} plots cumulative success rate against turn budget.
The ARGOS Agent resolves tasks substantially faster than w/o Strategy;
at turn~5 on Track~3, the gap is 60.8\% vs.\ 34.7\%, visually confirming the efficiency advantage that TWS quantifies.

\vspace{-2mm}
\paragraph{Results hold across seeds and surface variation.}
Across three seeds, ARGOS is stable (TWS std within $\pm$0.006; Sec.~B).
Under paraphrased witness responses, performance tracks witness \emph{uncertainty} rather than surface form: hedged phrasing lengthens dialogues and lowers TWS, while concise rewording matches the default templates.

\vspace{-2mm}
\section{Discussion and Conclusion}
\label{sec:conclusion}
\vspace{-3mm}

We presented ARGOS, a benchmark and agent for interactive multi-camera person search. Across four LLM backbones, only tool-augmented interaction grounded in the STTG reaches meaningful performance, and the benchmark remains far from solved.

\vspace{-3mm}
\paragraph{Limitations and broader impact.}
Three limitations merit discussion.
First, the witness simulator is deterministic by design, which isolates agent reasoning failures from witness noise. Real witnesses still exhibit memory errors and inconsistencies beyond our templates, and a stochastic or adversarial simulator is a natural next step.
Second, ARGOS currently covers two physical environments (factory, campus); scaling the STTG to highly disparate, city-scale camera networks with heterogeneous layouts and densities remains untested and may strain the construction pipeline.
Third, the technology underlying this work, namely identifying specific individuals from vague descriptions across camera networks, carries inherent risks of misuse in surveillance and privacy violation. We emphasize that ARGOS is designed as a research benchmark evaluated on fully consented, public datasets~\cite{woo2024mtmmc}; any deployment in real-world surveillance systems must comply with applicable privacy regulations and ethical guidelines.

\vspace{-3mm}
\paragraph{Future work.}
Promising directions include adversarial witness settings with misinformation and contradiction, multi-hop STTG reasoning over indirect camera paths, and transferring the ARGOS evaluation protocol to additional multi-camera datasets and environments.

\vspace{-5mm}
\subsubsection*{Acknowledgements.}
This work was supported by the Korea Institute of Science and Technology (KIST)
Institutional Program (Project No.~26K0030).

\bibliographystyle{splncs04}
\bibliography{main}

\end{document}